\documentclass{article}
\usepackage{spconf,amsmath,graphicx}
\usepackage{amssymb,amsfonts}
\graphicspath{{figs/}}
\usepackage{booktabs}
\usepackage{etoolbox}
\usepackage{siunitx}
\usepackage{textcomp}
\usepackage{xcolor}
\usepackage{multirow}
\usepackage[caption=false, font=footnotesize]{subfig}
\usepackage[export]{adjustbox}
\usepackage{enumitem}
\usepackage{cite}

\title{Ensemble of Models Trained by Key-based Transformed Images for Adversarially Robust Defense Against Black-box Attacks}
\name{MaungMaung AprilPyone and Hitoshi Kiya}
\address{Tokyo Metropolitan University, Tokyo, Japan}
\makeatletter
\renewcommand\section{\@startsection{section}{1}{\z@}
                      {0.5ex \@plus 0ex \@minus -2ex}
                      {0.5ex \@plus 0ex}
                      {\normalfont\Large\bfseries}}
\renewcommand\subsection{\@startsection{subsection}{2}{\z@}
                      {0.5ex \@plus 0ex \@minus -2ex}
                      {0.5ex \@plus 0ex}
                      {\normalfont\large\bfseries}}
\renewcommand\subsubsection{\@startsection{subsubsection}{3}{\z@}
                      {0.5ex \@plus 0ex \@minus -2ex}
                      {0.5ex \@plus 0ex}
                      {\normalfont\normalsize\bfseries}}
\def\@listi{\leftmargin\leftmargini
            \parsep 1.0pt
            \topsep 0.2\baselineskip \@minus 0.1\baselineskip
            \itemsep 1.0pt \relax}
\let\@listI\@listi
\makeatother
\begin{document}
%
\maketitle
\begin{abstract}
We propose a voting ensemble of models trained by using block-wise transformed images with secret keys for an adversarially robust defense. Key-based adversarial defenses were demonstrated to outperform state-of-the-art defenses against gradient-based (white-box) attacks. However, the key-based defenses are not effective enough against gradient-free (black-box) attacks without requiring any secret keys. Accordingly, we aim to enhance robustness against black-box attacks by using a voting ensemble of models. In the proposed ensemble, a number of models are trained by using images transformed with different keys and block sizes, and then a voting ensemble is applied to the models. In image classification experiments, the proposed defense is demonstrated to defend state-of-the-art attacks. The proposed defense achieves a clean accuracy of \SI{95.56}{\percent} and an attack success rate of less than \SI{9}{\percent} under attacks with a noise distance of 8/255 on the CIFAR-10 dataset.

\end{abstract}
\section{Introduction\label{sec:intro}}


Intentionally perturbed data points known as adversarial examples are indistinguishable from clean data points, but they cause deep neural networks (DNNs) erroneous predictions~\cite{2014-ICLR-Szegedy,2015-ICLR-Goodfellow}. The existence of such adversarial examples has been an alarming concern since DNNs are to be deployed in security-critical applications such as autonomous vehicles, healthcare, and finance. Therefore, a lot of effort has been put towards adversarial robustness.

Researchers have proposed various adversarial attacks and defenses.
However, most of the conventional defenses either reduce classification accuracy significantly (e.g., adversarial training~\cite{2018-ICLR-Madry}) or are completely broken~\cite{2020-Arxiv-Trammer,2018-ICML-Athalye}.
The research on adversarial robustness has entered into an “arms race” between attacks and defenses~\cite{2020-Arxiv-Trammer,2018-ICML-Uesato,2017-AISec-Carlini,2018-ICML-Athalye}.
Recent works~\cite{2018-ECCV-Taran,2020-ICIP-Maung, 2020-Arxiv-Maung} proposed defense methods with a secret key from a cryptographic point of view.
The main idea of these methods is to embed a secret key into the model structure with minimal impact on model performance.
Assuming the key stays secret, an attacker cannot compute any useful gradients on the model, which will render the existing gradient-based adversarial attack such as Projected Gradient Descent (PGD)~\cite{2018-ICLR-Madry} ineffective.
However, an attacker may perform gradient-free attacks (i.e., black-box attacks) without the secret key when the output of a model is available.

Therefore, in this work, we adapt the defense method in~\cite{2020-ICIP-Maung} and propose a voting ensemble to defend against black-box adversarial examples.
In the proposed method, a number of models are trained by using images transformed with different keys and block sizes. One of the models is used as a front-end model and outputs the probability of prediction to users, and a final class label is determined by using majority votes of all the models in the ensemble. We make the following contributions in this paper.
\begin{itemize}
\item We propose an ensemble of models protected by using secret keys for the first time.
\item We conduct three state-of-the-art black-box attacks and present the effectiveness of the proposed defense.
\end{itemize}
In image classification experiments, the proposed defense is confirmed not only to outperform the previous key-based adversarial defenses~\cite{2020-ICIP-Maung, 2018-ECCV-Taran}, but also to outperform state-of-the-art defenses~\cite{2020-ICLR-Wong, 2019-NIPS-Zhang}.

\section{Related Work}
\subsection{Adversarial Attacks}
Mainly, there are two types of attacks: white-box (complete knowledge of the model and its training data) and black-box (no knowledge).

\textbf{White-box:} Given an input image $x$ and a classifier $f(\cdot)$, an adversarial example $x'$ is generated such that $f(x') \neq y$, where $y$ is a true class. This is done by minimizing the perturbation $\delta$,
\begin{equation}
 \underset{\delta}{\text{minimize}} \left\rVert \delta \right\rVert_p, \;\;\text{s.t.}\;\; f_\theta(x + \delta) \neq y,
\end{equation}
or by maximizing the loss function,
\begin{equation}
 \underset{\delta \in \Delta}{\text{maximize}}\; \mathcal{L}(f_\theta(x + \delta), y).
\end{equation}
Usually, a typical threat model is bounded by an $\ell_p$ norm such that $\Delta = \{\delta: \left\rVert \delta \right\rVert_p \leq \epsilon \}$ for some perturbation distance $\epsilon > 0$.
Some of the most popular gradient-based (white-box) attacks are Fast Gradient Sign Method (FGSM)~\cite{2015-ICLR-Goodfellow}, Projected Gradient Descent (PGD)~\cite{2018-ICLR-Madry}, Carlini and Wagner (CW)~\cite{2017-SP-Carlini}, etc.

\textbf{Black-box:} There are also gradient-free (black-box) methods that estimate gradients such as~\cite{2018-ICML-Uesato,2019-NIPS-Cheng,2018-ICML-Ilyas}.
Another recent black-box attack, NATTACK, learns a probability distribution centered around the input such that a sample drawn from that distribution is likely an adversarial example~\cite{2019-ICML-Yandong}.
A recent score-based black-box attack based on a randomized search scheme is even more powerful than gradient-based white-box attacks~\cite{2020-ECCV-Maksym}.

\subsection{Adversarial Defenses}
The goal of a defense method is to make a model that is accurate not only for clean input but also for adversarial examples.
Many different approaches that try to achieve this goal, such as certified and provable defenses\cite{2019-NIPS-Salman, 2018-NIPS-Wong}, adversarial training~\cite{2019-NIPS-Shafahi,2020-ICLR-Wong,2018-ICLR-Madry}, preprocessing techniques~\cite{2018-ICLR-Guo,2018-ICLR-Buckman}, and detection algorithms~\cite{2017-ICLR-Metzen,2017-Arxiv-Feinman}.
All of the defense methods either reduce classification accuracy or are completely broken~\cite{2020-Arxiv-Trammer,2018-ICML-Athalye} and detection methods can be bypassed~\cite{2017-AISec-Carlini}.

Recently, a new line of adversarial defense methods was introduced by using a secret key as in cryptography~\cite{2018-ECCV-Taran,2020-ICIP-Maung}.
The work by~\cite{2018-ECCV-Taran} utilized pixel shuffling in a pixel-wise manner to input images by standard random permutation.
However, their work dropped accuracy drastically even on dataset like CIFAR-10.
The work by~\cite{2020-ICIP-Maung} proposed pixel shuffling in a block-wise manner to maintain high classification accuracy.
Although their work was confirmed to be effective against white-box attacks when the key is secret, black-box attacks that do not require the secret key to compute gradients are possible.
In this work, we adapt the work by~\cite{2020-ICIP-Maung} and propose a voting ensemble method to prevent from black-box attacks.

\begin{figure}[tbp]
  \centerline{\includegraphics[width=\linewidth]{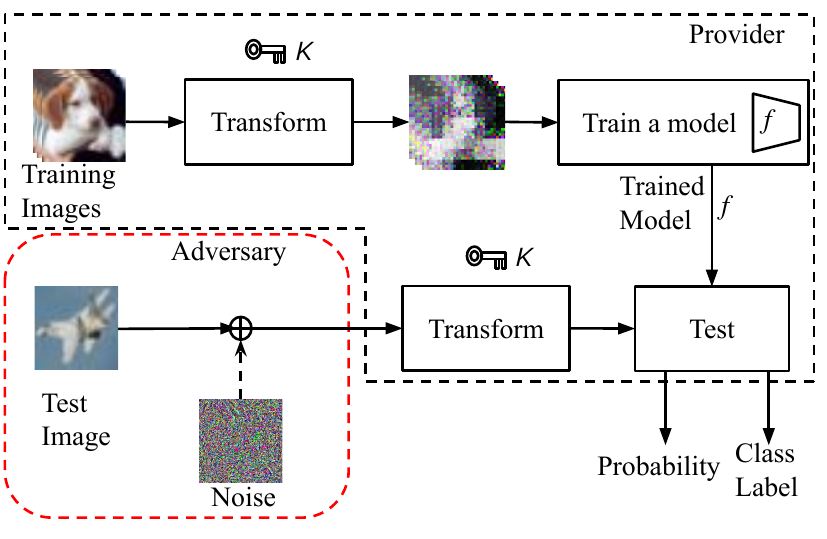}}
  \caption{Image classification with key-based transformation~\cite{2020-ICIP-Maung}.\label{fig:previous}}
\end{figure}

\section{Proposed Method}
\subsection{Overview}
Previous key-based adversarial defenses~\cite{2020-Arxiv-Maung, 2020-ICIP-Maung, 2018-ECCV-Taran} work well against white-box attacks by holding a secret key at a provider as shown in Fig.~\ref{fig:previous}. White-box attacks require correct gradients that cannot be computed without the secret key. Therefore, the previous works\cite{2020-Arxiv-Maung, 2020-ICIP-Maung, 2018-ECCV-Taran} are effective against white-box attacks. However, gradient-free (black-box) attacks that do not need to access the secret key can be still applied to the key-based defenses. Therefore, the proposed defense aims to extend the key-based defenses to defend against black-box attacks.

\begin{figure}[tbp]
  \centerline{\includegraphics[width=\linewidth]{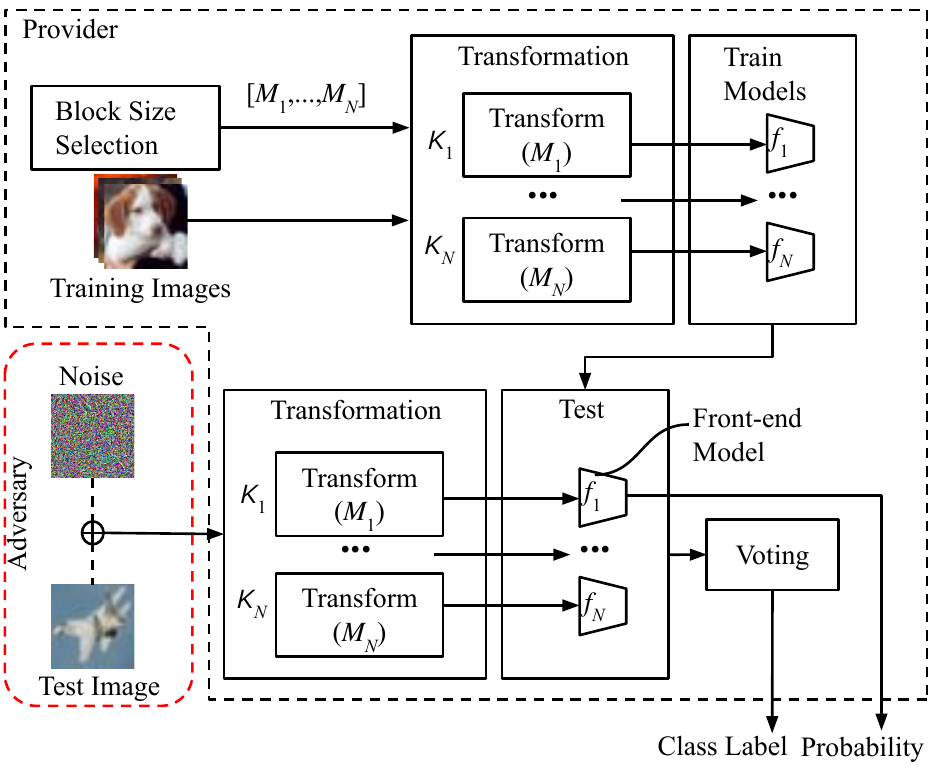}}
\caption{Image classification with proposed defense.\label{fig:overview}}
\end{figure}

The proposed method is an ensemble of models trained by using images transformed with different keys and block sizes. An overview of the proposed defense is shown in Fig.~\ref{fig:overview}. We first select block size $M_n \in\{M_1,\ldots,M_N\}$, and then $N$ models are trained by using images transformed with the selected $M_n$ and a key $K_n$.
One of the models in the ensemble is a front-end model (i.e., public-facing model) that outputs the probability of the prediction.
A final class label is determined by voting prediction results from all models.

\subsection{Block-wise Image Transformation with Secret Keys\label{sec:transformation}}
Models in the proposed ensemble are trained by using block-wise transformed images with secret keys as well as in~\cite{2020-ICIP-Maung}.
The following are steps for transforming images, where $c$, $w$ and $h$ denote the number of channels, width, and height of an image tensor $x \in {[0, 1]}^{c \times w \times h}$, and $N$ is the number of models.

\begin{enumerate}
  \item Divide $x$ into blocks with a size of $M_n \in\{M_1,\ldots,M_N\}$ such that $\{B_{(1,1)}, \ldots, B_{(\frac{w}{M_n}, \frac{h}{M_n})}\}$.
  \item Transform each block tensor $B_{(i, j)}$ into a vector $b_{(i,j)} = [b_{(i,j)}(1), \ldots, b_{(i,j)}(c \times M_n \times M_n)]$.
  \item Generate a random permutation vector $v$ with a key $K_n \in \{K_1,\ldots, K_N\}$, such that \\$[v_1, \dots, v_k, \dots, v_{k'}, \dots, v_{c \times M_n \times M_n}]$, where $v_k \neq v_{k'}$ if $k \neq k'$.
  \item Permutate every vector $b_{(i, j)}$ with $v$ as
    \begin{equation}
    b'_{(i, j)}(k) = b_{(i, j)}(v_k),
    \end{equation}
    to obtain a shuffled vector, $b'_{(i, j)}$ i.e., \\$[b'_{(i,j)}(1), \dots, b'_{(i,j)}(c \times M_n \times M_n)]$.
  \item Integrate the shuffled vectors to form a shuffled image tensor $x_n' \in {[0, 1]}^{c \times w \times h}$.
\end{enumerate}

The above steps are carried out for every model $f_n$, $n \in\{1,\ldots,N\}$.
Each model ($f_n$) in the ensemble is trained by using transformed images ($x_n'$).
Model $f_1$ is a front-end (public-facing) model which outputs a class probability to users.

\subsection{Selection of Block Size\label{sec:block-size}}
The previous works~\cite{2020-ICIP-Maung, 2020-Arxiv-Maung} confirmed that selecting a larger block size for block-wise image transformation provides better robustness against attacks, but a lower classification accuracy. In contrast, using a smaller block size achieves a higher classification accuracy, but lower robustness. From these results, we chose two block sizes for training models: a block size of $M_1 = 16$ for the frontend (public-facing) model and $M_{(2\text{ to } N)} = 2$ for the other models. The effectiveness of this selection will be demonstrated in an experiment.



\subsection{Ensemble}
Ensemble methods have been used as a technique applied to the improvement of model predictions in general~\cite{2015-IJCV-Russakovsky}. There are many forms of the ensemble:
voting, bagging, boosting, and stacking. In this work, we utilize voting where the output class label is determined by the majority of votes. The proposed voting ensemble consists of two steps:
\begin{enumerate}
  \item Each model computes an output probability vector $z_n = [z_{n} (1),\ldots,z_{n} (L)]$, where $L$ is the number of class labels. 
  \item The final prediction $y_{\text{final}}$ is determined by using the majority of the votes from the models, i.e.,
    \begin{equation}
      y_{\text{final}} = \text{mode}(\text{argmax}(z_1), \ldots, \text{argmax}(z_N)).
    \end{equation}
\end{enumerate}


\robustify\bfseries
\sisetup{table-parse-only,detect-weight=true,detect-inline-weight=text,round-mode=places,round-precision=2}
\begin{table*}[!ht]
  \caption{Clean ACC (\SI{}{\percent}) and ASR (\SI{}{\percent}) of proposed ensemble under three black-box attacks\label{tab:results}}
  \centering
  \begin{tabular}{l|S|SSS}
  \toprule
  & {(ACC)} & \multicolumn{3}{c}{(ASR)}\\
  {Model} & {Clean} & {SPSA~\cite{2018-ICML-Uesato}} & {NATTACK~\cite{2019-ICML-Yandong}} & {SQUARE~\cite{2020-ECCV-Maksym}}\\
  \midrule
  Proposed ($M_{(1\text{ to } N)} = 2, N = 9$) & 95.38 & 58.33 & 9.21 & 11.32\\
  Proposed ($M_{(1\text{ to } N)} = 4, N = 9$) & 93.17 & 34.29 & 8.98 & 10.38\\
  Proposed ($M_{(1\text{ to } N)} = 8, N = 9$) & 88.47 & 44.52 & 8.90 & 16.05\\
  Proposed ($M_{(1\text{ to } N)} = 16, N = 9$) & 79.63 & 25.71 & 6.18 & 16.49\\
  \midrule
  Proposed ($M_1 = 16, M_{(2\text{ to } N)} = 2, N = 9$) & \bfseries \num{95.56} & \bfseries \num{8.90} & \bfseries \num{1.73} & \bfseries \num{0.82}\\
  Proposed ($M_1 = 8, M_{(2\text{ to } N)} = 2, N = 9$) & 95.31 & 17.39 & 1.46 & 3.46\\
  Proposed ($M_1 = 4, M_{(2\text{ to } N)} = 2, N = 9$) & 95.30 & 30.95 & 6.45 & 7.12\\
  \midrule
  Baseline & 95.45  & 100.0 & 99.64 & 99.19\\
  \bottomrule
  \end{tabular}
\end{table*}
\robustify\bfseries
\sisetup{table-parse-only,detect-weight=true,detect-inline-weight=text,round-mode=places,round-precision=2}
\begin{table*}[!ht]
  \caption{ACC (\SI{}{\percent}) of proposed ensemble with various enemble sizes\label{tab:size}}
  \centering
  \begin{tabular}{lSSSS}
  \toprule
  {Model} & {Clean} & {SPSA~\cite{2018-ICML-Uesato}} & {NATTACK~\cite{2019-ICML-Yandong}} & {SQUARE~\cite{2020-ECCV-Maksym}}\\
  \midrule
  Proposed ($M_1 = 16, M_{(2\text{ to } N)} = 2, N = 3$) & 94.46 & 83.69 & 92.14 & 90.59\\
  Proposed ($M_1 = 16, M_{(2\text{ to } N)} = 2, N = 5$) & 95.28 & 86.44 & 93.89 & 92.86\\
  Proposed ($M_1 = 16, M_{(2\text{ to } N)} = 2, N = 7$) & 95.36 & \bfseries \num{87.13} & 94.20 & 93.24\\
  Proposed ($M_1 = 16, M_{(2\text{ to } N)} = 2, N = 9$) & \bfseries \num{95.56} & 86.97 & \bfseries \num{94.59} & \bfseries \num{93.56}\\
  \midrule
  Baseline & 95.45  & 0.02 & 0.32 & 0.76\\
  \bottomrule
  \end{tabular}
\end{table*}

\subsection{Threat Model}
To evaluate a defense method, precisely defining threat models is necessary.
A threat model includes a set of assumptions such as an adversary's goals, knowledge, and capabilities~\cite{2019-Arxiv-Carlini}.
We consider untargeted attacks where the goal of the adversary is to reduce the classification accuracy of the model.
In~\cite{2020-ICIP-Maung}, it is confirmed that white-box attacks such as PGD~\cite{2018-ICLR-Madry} are not effective when the key is secret because correct gradients cannot be computed without the key.
Therefore, in this work, we focus on black-box attacks, specifically, SPSA~\cite{2018-ICML-Uesato}, NATTACK~\cite{2019-ICML-Yandong}, and SQUARE~\cite{2020-ECCV-Maksym} attacks.
The adversary performs untargeted evasion attacks (i.e., test time attacks) in which small changes under $\ell_\infty$ metric change the true class of the input.
The adversary's capability is to modify the test image where the noise distance $\epsilon$ is $8/255$.

\section{Experiments}
\subsection{Set-up}
We used the CIFAR-10~\cite{2009-Report-Krizhevsky} dataset with a batch size of 128 and live augmentation (random cropping with padding of 4 and random horizontal flip) on the training set.
CIFAR-10 consists of 60,000 color images (dimension of $32 \times 32 \times 3$) with 10 classes (6000 images for each class) where 50,000 images are for training and 10,000 for testing.

We utilized deep residual networks~\cite{2016-CVPR-He} with 18 layers (ResNet18) for the CIFAR-10 dataset and trained for $200$ epochs with efficient training techniques from the DAWNBench top submissions: cyclic learning rates and mixed-precision training.
The parameters of the stochastic gradient descent (SGD) optimizer were a momentum of $0.9$, weight decay of $0.0005$, and a maximum learning rate of $0.2$.

Three black-box attacks: SPSA~\cite{2018-ICML-Uesato}, NATTACK~\cite{2019-ICML-Yandong}, SQUARE~\cite{2020-ECCV-Maksym} were deployed to evaluate the proposed defense.
NATTACK was configured with a population size of 300, a sigma value of 0.1, a learning rate of 0.02, and 100 iterations.
SPSA was set up with a delta value of 0.01, a learning rate of 0.01, a batch size of 256, and 100 maximum iterations.
We used 2000 queries for SQUARE attack.
The noise distance $\epsilon$ was 8/255 for all attacks.

To evaluate the proposed defense, we used accuracy (ACC) over the whole test dataset (10,000 images) and attack success rate (ASR) over randomly selected 1000 images that are correctly classified by the ensemble.
The goal of a defense method is to make a model that has both a high ACC value and a low ASR one.

\subsection{Evaluation of Proposed Defense}
Table~\ref{tab:results} summarizes the performances of the proposed defense in terms of ACC and ASR values under the use of various block sizes used in image transformation, where $M_{(1\text{ to } N)}$ indicates $M_1 = M_2 = \cdots = M_N$, and $M_{(2\text{ to } N)}$ means $M_2 = M_3 = \cdots = M_N$, and $N$ is the number of models.
When the same block size as $M_{(1\text{ to } N)}$ was applied to all models,  the proposed ensemble with $M_{(1\text{ to } N)} = 2$ achieved almost the same accuracy as the baseline model (without any defense) in terms of clean accuracy. However, as demonstrated in~\cite{2020-ICIP-Maung, 2020-Arxiv-Maung}, using a smaller block size for the block-wise transformation achieves a higher classification accuracy on clean images, but the robustness against adversarial examples is lower (a higher ASR value).
Therefore, to achieve both a higher ACC value on clean images and a low ASR even under attacks, we selected two block sizes for training models such as $M_1 = 16$ and $M_{(2\text{ to } N)} = 2$. From Table~\ref{tab:results}, the proposed defense with two block sizes allowed us not only to maintain high ACC values but also to achieve lower ASR ones. In particular, The ensemble with $M_1 = 16$ and $M_{(2\text{ to } N)} = 2$ outperformed the other ensembles under the attacks.

\subsection{Selection of the Number of Models}
The experiment in Table~\ref{tab:results} was carried out with an ensemble size of $N = 9$.
In contrast, Table~\ref{tab:size} shows the performance of the proposed defense under the use of variable ensemble sizes (i.e., $N \in\{3, 5, 7, 9\}$) in terms of ACC values for both clean images and adversarial examples.
From the table, the proposed defense was demonstrated to be able to maintain high ACC values even under variable ensemble sizes. In particular, the ensemble with $N = 9$ achieved the highest clean accuracy.

\subsection{Comparison with State-of-the-art Defenses}
We compared the proposed ensemble with state-of-theart defenses: fast adversarial training (Fast AT)~\cite{2020-ICLR-Wong}, feature scattering approach (FS)~\cite{2019-NIPS-Zhang}, key-based standard random permutation (SRP)~\cite{2018-ECCV-Taran} and block-wise shuffling (Key-based $M = 4$)~\cite{2020-ICIP-Maung} in terms of clean ACC and ASR.\@
The ASR of Fast AT for NATTACK was $\approx$ \SI{40}{\percent} and that of the other two attacks was $\approx$ \SI{32}{\percent}.
The ASR of FS was lower than that of Fast AT (\SI{20.48}{\percent} for NATTACK and $\approx$ \SI{30}{\percent} for the other two attacks).
The models, Key-based ($M = 4$) and SRP were completely defeated by black-box attacks (i.e.,  the ARS was almost \SI{100}{\percent}).
In contrast, the ASR of the proposed ensemble was lower than \SI{9}{\percent} for SPSA and less than \SI{2}{\percent} for the other two attacks.
Therefore, the proposed defense outperformed the state-of-the-art defenses under all the three black-box attacks.

\robustify\bfseries
\sisetup{table-parse-only,detect-weight=true,detect-inline-weight=text,round-mode=places,round-precision=2}
\begin{table}[tbp]
  \caption{Comparison with state-of-the-art defenses in terms of clean ACC (\SI{}{\percent}) and ASR (\SI{}{\percent})\label{tab:comparison}}
  \centering
  \resizebox{\columnwidth}{!}{%
  \begin{tabular}{l|S|SSS}
  \toprule
  & {(ACC)} & \multicolumn{3}{c}{(ASR)}\\
  {Model} & {Clean} & {SPSA~\cite{2018-ICML-Uesato}} & {NATTACK~\cite{2019-ICML-Yandong}} & {SQUARE~\cite{2020-ECCV-Maksym}}\\
  \midrule
  Proposed$^{\dagger}$  & \bfseries \num{95.56} & \bfseries \num{8.90} & \bfseries \num{1.73} & \bfseries \num{0.82}\\
  \midrule
  Baseline & 95.45  & 100.0 & 99.64 & 99.19\\
  \midrule
  Key-based ($M = 4$)~\cite{2020-ICIP-Maung} & 91.84 & 99.71 & 99.90 & 98.86\\
  Fast AT~\cite{2020-ICLR-Wong} & 83.80 & 33.47 & 40.06 & 33.50\\
  FS~\cite{2019-NIPS-Zhang} & 89.98 & 29.47 & 20.48 & 31.87\\
  SRP~\cite{2018-ECCV-Taran} & 65.16 & 99.70 & 99.40 & 98.80\\
  \bottomrule
  \multicolumn{5}{l}{$^{\dagger}$($M_1 = 16, M_{(2\text{ to } N)} = 2, N = 9$)}
  \end{tabular}
}
\end{table}

\section{Conclusion}

In this paper, we proposed a new adversarial defense that is a voting ensemble of key-protected models to overcome black-box attacks for the first time.
Models in the ensemble were trained by using images transformed with different keys.
We evaluated the proposed defense under three different state-of-the-art black box attacks.
The results showed that clean accuracy of the proposed defense achieved more than \SI{95}{\percent} and the attack success rate was less than \SI{9}{\percent} for the three black-box attacks with a noise distance of 8/255 on the CIFAR-10 dataset.
Comparing with state-of-the-art defenses, the proposed ensemble outperformed both the previous key-based defenses and the conventional adversarial defenses.
As for future work, we shall optimize the ensemble to further improve the performance of the proposed defense.





\begin{small}
\bibliographystyle{IEEEbib}
\bibliography{refs}
\end{small}

\end{document}